\documentclass[sigconf,screen]{acmart}

\AtBeginDocument{%
  \providecommand\BibTeX{{%
    \normalfont B\kern-0.5em{\scshape i\kern-0.25em b}\kern-0.8em\TeX}}}

\setcopyright{acmcopyright}
\copyrightyear{2022}
\acmYear{2022}
\acmDOI{10.1145/3503161.3548086}

\acmConference[MM '22]{Proceedings of the 30th ACM International Conference on Multimedia}{October 10--14, 2022}{Lisboa, Portugal}
\acmBooktitle{Proceedings of the 30th ACM International Conference on Multimedia (MM '22), October 10--14, 2022, Lisboa, Portugal}
\acmPrice{15.00}
\acmISBN{978-1-4503-9203-7/22/10}

\acmSubmissionID{1398}

\usepackage{bm}
\usepackage{subcaption}
\usepackage{tikz}
\usepackage{pgfplots}
\usepackage{multirow}
\usepackage{balance}

\pgfplotsset{compat=newest}

\definecolor{mygray}{gray}{.9}
\definecolor{sota_blue}{HTML}{0071bc}
\makeatletter
\newcommand{\thickhline}{%
	\noalign {\ifnum 0=`}\fi \hrule height 1pt
	\futurelet \reserved@a \@xhline
}
\makeatother
\newcommand{\tablestyle}[2]{\setlength{\tabcolsep}{#1}\renewcommand{\arraystretch}{#2}\centering\footnotesize}

\begin{document}

% \title{P2PM: A Simple yet Effective One-Stage Approach for \\
% Panoptic Narrative Grounding}
\title{PPMN: Pixel-Phrase Matching Network for One-Stage Panoptic Narrative Grounding}

% \affiliation{%
\author{Zihan Ding}
\authornote{Equal contribution.}
\email{dingzihan737@gmail.com}
\affiliation{
    \institution{Institute of Artificial Intelligence, Beihang University}
    \institution{Hangzhou Innovation Institute, Beihang University}
    \country{}
}

\author{Zi-han Ding}
\email{zihanding819@gmail.com}
\authornotemark[1]
\affiliation{
    \institution{Institute of Artificial Intelligence, Beihang University}
    \institution{Hangzhou Innovation Institute, Beihang University}
    \country{}
}

\author{Tianrui Hui}
\email{huitianrui@gmail.com}
\authornote{Corresponding author.}
\affiliation{
    \institution{Institute of Information Engineering, Chinese Academy of Sciences}
    \institution{School of Cyber Security, University of Chinese Academy of Sciences
    }
    \country{}
}

\author{Junshi Huang}
\affiliation{
    \institution{Meituan}
    \country{}
}

\author{Xiaoming Wei}
\affiliation{
    \institution{Meituan}
    \country{}
}

\author{Xiaolin Wei}
\affiliation{
    \institution{Meituan}
    \country{}
}

\author{Si Liu}
\affiliation{
    \institution{Institute of Artificial Intelligence, Beihang University}
    \institution{Hangzhou Innovation Institute, Beihang University}
    \country{}
}

\renewcommand{\shortauthors}{Zihan Ding et al.}

\begin{abstract}
Panoptic Narrative Grounding (PNG) is an emerging task whose goal is to segment visual objects of \textit{things} and \textit{stuff} categories described by dense narrative captions of a still image.
The previous two-stage approach first extracts segmentation region proposals by an off-the-shelf panoptic segmentation model, then conducts coarse region-phrase matching to ground the candidate regions for each noun phrase.
However, the two-stage pipeline usually suffers from the performance limitation of low-quality proposals in the first stage and the loss of spatial details caused by region feature pooling, as well as complicated strategies designed for \textit{things} and \textit{stuff} categories separately.
To alleviate these drawbacks, we propose a one-stage end-to-end Pixel-Phrase Matching Network (PPMN), which directly matches each phrase to its corresponding pixels instead of region proposals and outputs panoptic segmentation by simple combination.
Thus, our model can exploit sufficient and finer cross-modal semantic correspondence from the supervision of densely annotated pixel-phrase pairs rather than sparse region-phrase pairs.
In addition, we also propose a Language-Compatible Pixel Aggregation (LCPA) module to further enhance the discriminative ability of phrase features through multi-round refinement, which selects the most compatible pixels for each phrase to adaptively aggregate the corresponding visual context.
Extensive experiments show that our method achieves new state-of-the-art performance on the PNG benchmark with 4.0 absolute Average Recall gains\footnote[1]{https://github.com/dzh19990407/PPMN}.
\end{abstract}




\begin{teaserfigure}
  \centering
  \medskip
  \includegraphics[width=0.9\textwidth]{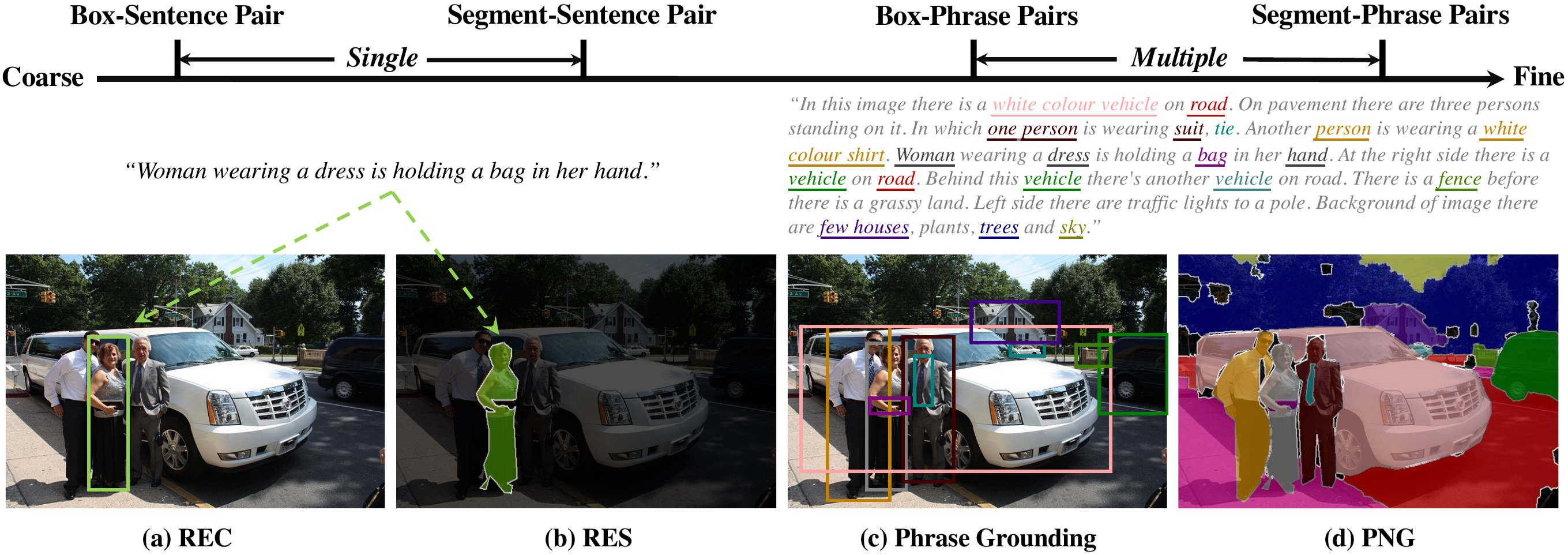}
  \caption{Illustration of different visual grounding tasks. (a) Referring Expression Comprehension (REC) aims to locate the object referred by a sentence. (b) Referring Expression Segmentation (RES) aims to segment the object referred by a sentence. (c) Phrase Grounding aims to locate multiple objects referred by noun phrases. (d) Panoptic Narrative Grounding (PNG) aims to generate a panoptic segmentation according to dense narrative captions. For better viewing of all figures in this paper, please see original zoomed-in color pdf file.}
  \label{fig:teaser}
\end{teaserfigure}

\maketitle

\section{Introduction}

Panoptic Narrative Grounding (PNG)~\cite{gonzalez2021panoptic}, which aims to segment visual objects of both~\textit{things} and~\textit{stuff} (\textit{i.e.}, panoptic) categories in an image grounded by dense narrative captions, is a newly proposed and more general formulation of the natural language visual grounding problem.
In Figure~\ref{fig:teaser} we illustrate the difference between PNG and traditional visual grounding tasks.
Referring expression comprehension and segmentation~\cite{luo2020cascade,cheng2021exploring,li2021bottom,liao2022progressive,liu2021cross,hui2020linguistic} ground the subject of a single sentence in the form of a single box and a single segment respectively, while phrase grounding~\cite{ye2021one,jing2020visual} extends them by locating multiple objects grounded by multiple noun phrases in the caption.
Among these grounding tasks, PNG achieves the finest cross-modal alignment between multiple noun phrases and multiple segments. 
Compared to Panoptic Segmentation~\cite{kirillov2019panoptic}, PNG also possesses more flexibility as it requires the model to produce a panoptic segmentation according to free-form narrative captions instead of a fixed set of pre-determined object categories.
With spatially finer and denser cross-modal alignment between noun phrases and segmentation masks, PNG is able to benefit many downstream applications including text-driven image manipulation~\cite{zhang2021text}, visual question answering~\cite{antol2015vqa}, and intelligent robots~\cite{zhang2020language},~\textit{etc}.

\begin{figure}[t]
  \centering
  \includegraphics[width=\linewidth]{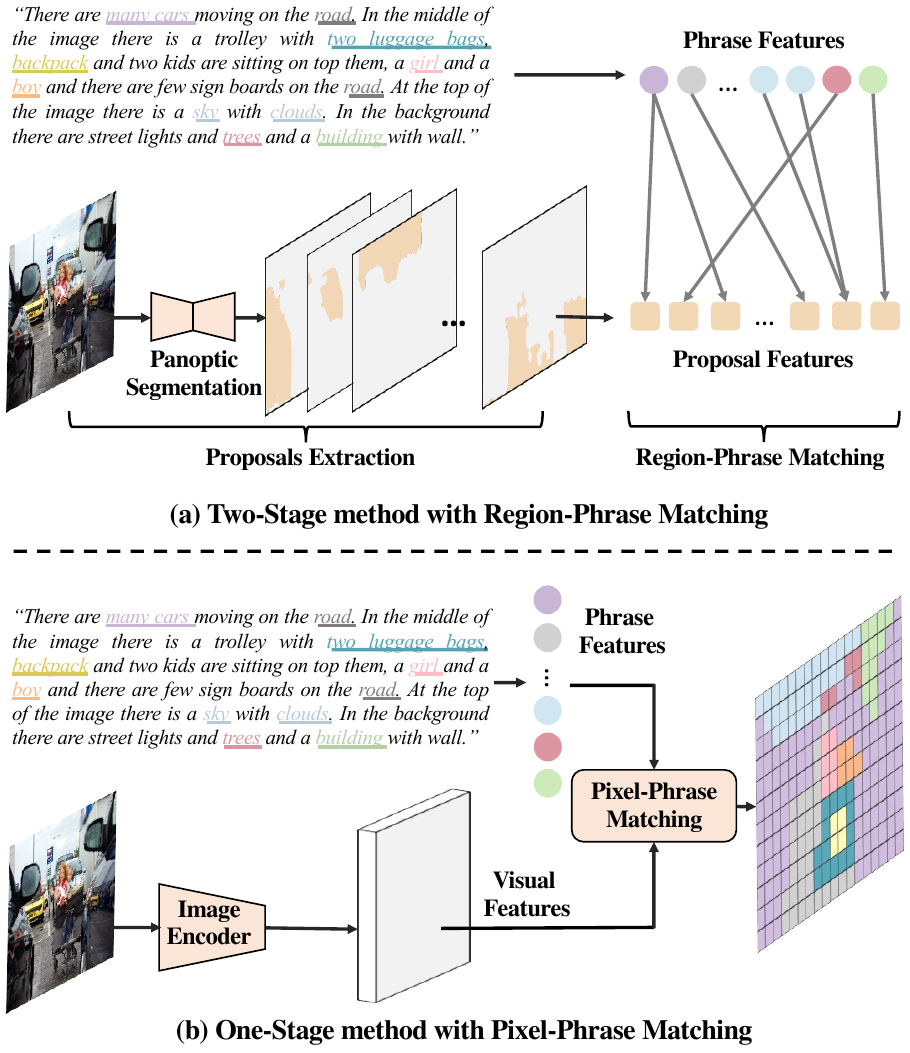}
  \caption{Comparison between the two-stage baseline~\cite{gonzalez2021panoptic} and our one-stage Pixel-Phrase Matching Network (PPMN). (a) The two-stage baseline first extracts segmentation proposals using an off-the-shelf panoptic segmentation model. Then, it conducts coarse Region-Phrase Matching to select grounding results for each noun phrase. (b) Our one-stage PPMN conducts finer pixel-phrase matching by calculating the matching scores between features of all pixels and all phrases to directly generate panoptic segmentation.}
  \label{fig:motivation}
  \vspace{-10pt}
\end{figure}

Along with the PNG benchmark, Gonz{\'a}lez~\textit{et al.}~\cite{gonzalez2021panoptic} also provide a strong~\textbf{two-stage} baseline upon the Cross-Modality Relevance (CMR)~\cite{zheng2020cross} model (Figure~\ref{fig:motivation}(a)).
Concretely, they first extract the region feature of each segmentation proposal from an off-the-shelf panoptic segmentation model~\cite{kirillov2019panopticfpn}.
Then, they match (\textit{i.e.}, ground) candidate segmentation regions conditioning on the affinity matrix calculated between features of all region proposals and noun phrases.
Although this baseline has achieved good performance, some limitations may still exist:
1) The matching performance in the second stage is restricted by the quality of segmentation proposals and the recall rate of the panoptic segmentation model in the first stage.
2) Region feature of each segmentation proposal is pooled into a single vector, in which detailed spatial information may be lost, incurring a coarse cross-modal matching process.
3) The two-stage pipeline is complicated and involves many manually designed rules, such as different feature extraction strategies for~\textit{things} and~\textit{stuff} categories, and the extra post-processing for plurals noun phrases in the inference phase.

In order to alleviate the above problems, we propose a simple yet effective ~\textbf{one-stage} approach for PNG, which enjoys end-to-end optimization.
Specifically, instead of conducting indirect and coarse Region-Phrase Matching with an off-the-shelf panoptic segmentation model, our one-stage approach is realized via a direct and fine Pixel-Phrase Matching Network (PPMN).
As illustrated in Figure~\ref{fig:motivation}(b), our PPMN matches each phrase to its corresponding pixels by directly calculating the matching matrix between features of all phrases and all pixels, so that each phrase can obtain a response map derived from the matching matrix.
The panoptic segmentation is then generated by simply combining the results of all phrases.
It is interesting to note that multiple phrases can correspond to the same ground-truth mask in the task setting of PNG.
Therefore, each pixel is actually classified to multiple labels (\textit{i.e.}, phrases) when applying binary cross-entropy loss on all response maps, where responses on positive labels are forced to be higher and otherwise on negative ones.
Overall, the merits of our PPMN contain three aspects:
1) One-stage framework can exploit sufficient cross-modal semantic correspondence from the supervision of densely annotated pixel-phrase pairs rather than sparse region-phrase pairs.
2) By avoiding proposal pooling, detailed spatial information can be preserved as well for more accurate and complete segmentation results.
3) Our PPMN can better tap the category priors contained intrinsically in the natural language to distinguish \textit{things}/\textit{stuff} and \textit{singulars}/\textit{plurals}, which removes complicated post-processing and largely simplifies the whole pipeline.

To further enhance the discriminative ability of each phrase feature, we propose a Language-Compatible Pixel Aggregation (LCPA) module.
In detail, we select the most compatible pixels for each phrase according to matching scores and adaptively aggregate their features via a multi-head cross-modal attention mechanism.
By this means, each phrase is aware of the corresponding visual contextual information.
Our LCPA module is applied for multiple rounds to gradually refine phrase features with adaptive visual clues for more accurate pixel-phrase matching.

The main contributions of our paper are summarized as follows:
1) We propose a novel one-stage end-to-end method termed Pixel-Phrase Matching Network (PPMN) for the emerging PNG task, where each phrase is directly matched with its corresponding pixels instead of pre-generated region proposals to produce finer panoptic segmentation.
2) For more accurate pixel-phrase matching, we also propose a Language-Compatible Pixel Aggregation (LCPA) module to gradually enhance the discriminative ability of phrase features with adaptive visual clues.
3) Extensive experiments show that our one-stage method outperforms the previous two-stage baseline on the PNG benchmark with a significant gain of 4.0 overall Average Recall.
Our code will be made publicly available to facilitate further research in this field.

\section{Related Work}

\subsection{Panoptic Segmentation}

Panoptic segmentation~\cite{kirillov2019panoptic}, which aims to assign a semantic label and an instance id to each pixel, has received much attention recently. 
Mainstream methods~\cite{kirillov2019panopticfpn,xiong2019upsnet,cheng2020panoptic,li2020unifying,tian2022instance} are usually composed of two specifically designed branches for segmenting foreground ~\textit{things} (\textit{i.e.,} instance segmentation) and background~\textit{stuff} (\textit{i.e.,} semantic segmentation) respectively.
For example, Kirillov~\textit{et al.}~\cite{kirillov2019panopticfpn} propose to extend Mask R-CNN~\cite{he2017mask} with a semantic segmentation branch using a shared FPN~\cite{lin2017feature} backbone.
With the rise of Transformer~\cite{vaswani2017attention}, some methods \cite{carion2020end, wang2021max, li2021panoptic,cheng2021per,zhang2021k} have emerged with an end-to-end set prediction objective, and generate panoptic masks through attention blocks. 
Differently, PNG aims to generate the panoptic segmentation for a still image according to dense narrative captions.
To tackle this task, we propose a simple yet effective one-stage method that directly matches each phrase with its corresponding pixels to segment both~\textit{things} and~\textit{stuff} categories in a unified way.

\begin{figure*}[t]
  \centering
  \includegraphics[width=\linewidth]{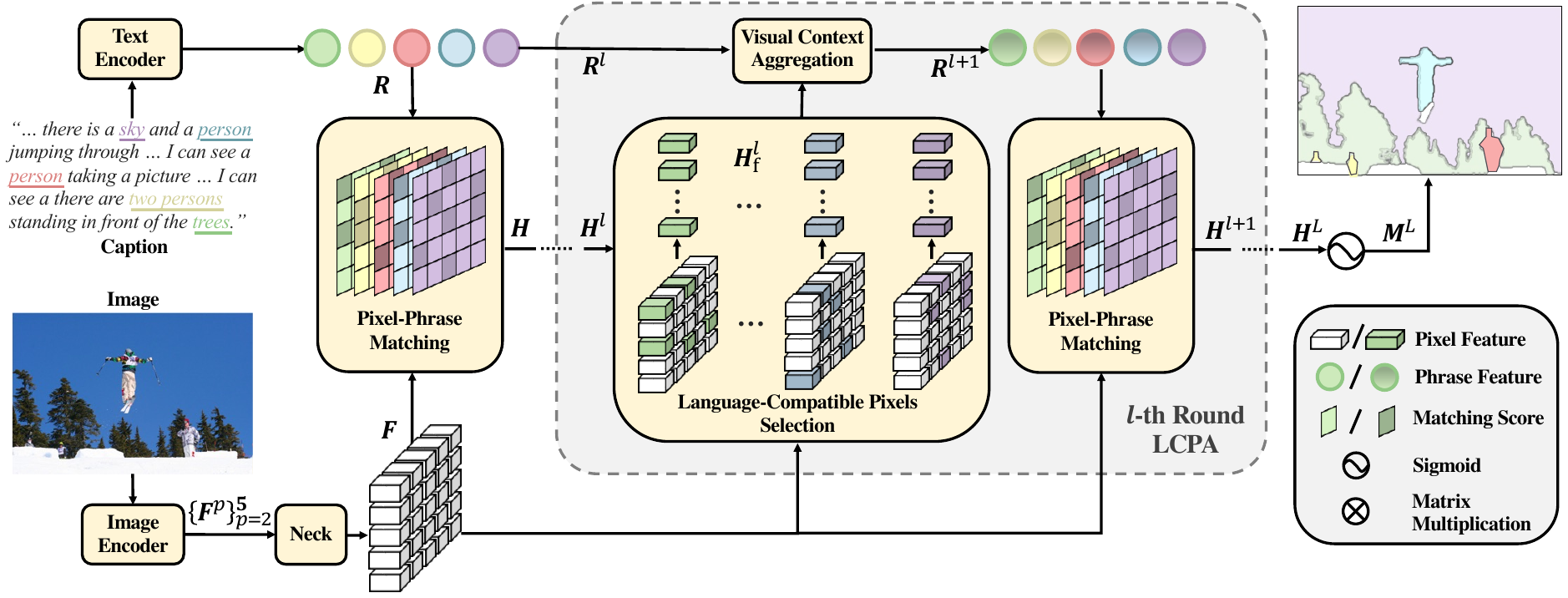}
  \caption{Overview of our Pixel-Phrase Matching Network (PPMN). We use an image encoder to extract multi-scale image features ${\{\bm{F}^p\}}_{p=2}^5$ and aggregate them into a single feature map $\bm{F}$.
  As for the linguistic modality, we use a text encoder to extract noun phrase features $\bm{R}$.
  We obtain the raw matching maps $\bm{H}$ via Pixel-Phrase Matching. 
  To elaborate on the LCPA approach, we take the $l$-th round as an example. 
  Firstly, we use $\bm{H}^l$ to select language-compatible pixels $\bm{H}_{\text{f}}^l$ for each noun phrase from $\bm{F}$.
  After, we aggregate the specific visual context using each phrase feature in $\bm{R}^l$ as the query to get the refined phrase features $\bm{R}^{l+1}$. 
  And we conduct Pixel-Phrase Matching to obtain the refined matching maps $\bm{H}^{l+1}$.
  After multiple rounds of refinement, we apply sigmoid on $\bm{H}^{L}$ from the last round to get response maps $\bm{M}^L$, from which we generate panoptic segmentation.}
  \label{fig:framework}
\end{figure*}

\subsection{Referring Expression Comprehension}

Referring Expression Comprehension (REC) aims to predict the bounding box of the object referred by the referring expression. 
Existing methods can be roughly divided into two categories: two-stage methods~\cite{yu2018mattnet,liu2019improving,zhang2018grounding,yang2019dynamic,wang2019neighbourhood,lu2019vilbert,su2019vl,chen2020uniter,jing2020visual} and one-stage methods~\cite{yang2019fast,yang2020improving,yang2020propagating,liao2020real,deng2021transvg}. 
Conventional two-stage methods follow the propose-and-match paradigm and design sophisticated ways of cross-modal interaction for accurate vision-language alignment (\textit{e.g.,} fine-grained context modeling~\cite{yu2018mattnet,liu2019improving,zhang2018grounding} and graph attention~\cite{yang2019dynamic,wang2019neighbourhood,jing2020visual}). 
Recently, some methods~\cite{lu2019vilbert,su2019vl,chen2020uniter} leverage the powerful modeling ability of BERT~\cite{devlin2018bert} to learn better vision-language representations on large-scale image-text datasets. By introducing language into one-stage detectors~\cite{redmon2016you,zhou2019objects}, one-stage approaches~\cite{yang2019fast,yang2020improving,yang2020propagating,liao2020real} are gradually taking over the mainstream as they achieve a good balance between effectiveness and efficiency.
Moreover, TransVG~\cite{deng2021transvg} take a further step for directly regressing the box coordinates while using Transformers~\cite{su2019vl} to conduct cross-modal alignment.

\subsection{Referring Expression Segmentation}
Referring expression segmentation (RES) requires models to predict foreground pixels for the object described by the input referring expression.
Hu~\textit{et al.}~\cite{hu2016segmentation} first propose an one-stage framework for RES, where they use FCN~\cite{long2015fully} and LSTM~\cite{hochreiter1997long} to extract visual feature maps and sentence features respectively.
Then, they fuse them by concatenation to form the cross-modal features, on which they apply deconvolution layers to generate the segmentation mask.
To exploit different types of informative words in the expression and accurately align the two modalities, CMPC~\cite{huang2020referring} first perceives all possible entities under the guidance of \textit{entity} and \textit{attribute} clues of the input expression and then utilizes \textit{relation} words to filter out irrelevant ones.
Motivated by the powerful ability of Transformers~\cite{vaswani2017attention} for capturing long-range dependencies, some methods~\cite{ding2021vision,li2021referring,jiao2021two,feng2021encoder,yang2021lavt} design complex cross-attention mechanisms to model the semantic relationship between vision and language modalities.
Moreover, RES can be applied on consecutive video frames for segmenting the queried actors~\cite{hui2021collaborative,ding2021progressive,ding2022language}.
Differently, PNG grounds multi noun phrases in the narrative caption for panoptic categories, and in this paper we formulate PNG as a direct pixel-phrase matching process to fully mine the cross-modal correspondence from densely annotated pixel-phrase pairs.

\subsection{Phrase Grounding}

Phase grounding aims to localize multiple regions with bounding boxes in an image referred by noun phrases in natural language descriptions. 
Early methods~\cite{karpathy2014deep,wang2018learning,plummer2018conditional,akbari2019multi} used to first extract region and phrase embeddings independently, and then learn the semantic correspondence between phrase-region pairs in a shared embedding subspace via manually designed loss functions.
In order to exploit visual and textual context, some methods begin to adopt different ways of cross-modal interaction~\cite{bajaj2019g3raphground,dogan2019neural,liu2020learning,mu2021disentangled,yu2020cross}.
Mu~\textit{et al.}~\cite{mu2021disentangled} propose to distinguish various contexts via motif-aware graph learning, which can achieve fine-grained contextual fusion.
Yu~\textit{et al.}~\cite{yu2020cross} explore how to capture omni-range dependencies through both multi-level and multi-modal interaction.
Moreover, recent research shows that phrase grounding can benefit a lot from the large-scale vision-language pretraining~\cite{kamath2021mdetr,li2021grounded}.
Compared with phrase grounding, PNG is spatially finer and more general as it includes segmentation annotations of all panoptic categories (\textit{i.e.,} \textit{things} and \textit{stuff}).
In this paper, we propose a one-stage Pixel-Phrase Matching Network (PPMN) to better mine the fine-grained semantic richness of a visual scene for the PNG task.

\section{Method}

In this section, we first introduce the feature extraction processes for visual and linguistic modalities (\S\ref{sec:fe}). 
Then, we describe how the panoptic narrative grounding task can be formulated as a pixel-phrase matching problem (\S\ref{sec:ppm}).
Finally, we introduce the proposed Language-Compatible Pixel Aggregation (LCPA) module which enhances the discriminative ability of phrase features by aggregating the corresponding visual context for each noun phrase to achieve accurate pixel-phrase matching (\S\ref{sec:lcpa}).
The overall pipeline of our Pixel-Phrase Matching Network (PPMN) is illustrated in Figure~\ref{fig:framework}.

\subsection{Feature Extraction}\label{sec:fe}

For the visual modality, we use FPN~\cite{lin2017feature} with a ResNet-101~\cite{he2016deep} backbone as the image encoder to extract multi-scale feature maps $\bm{F}^{p}\in \mathbb{R}^{H^{p}\times W^{p}\times C_{\text{v}}}, p\in \{2,3,4,5\}$, where $H^{p}=\frac{H^0}{2^{p}}$, $W^{p}=\frac{W^0}{2^{p}}$, and $C_{\text{v}}$ are the height, width, and channel number of the $p$-th visual features respectively, $H^0$ and $W^0$ are the original scale of the input image. 
To enhance the positional information, we add the sinusoids positional encoding~\cite{vaswani2017attention} with $\bm{F}^{5}$. Next, we send $\{\bm{F}^p\}_{p=2}^{5}$ into the semantic FPN neck~\cite{kirillov2019panopticfpn} to obtain the final visual feature map $\bm{F}\in \mathbb{R}^{H\times W\times C_{\text{v}}}$ with strong semantic representation 
and low-level local details, where $H=\frac{H^0}{8}$ and $W=\frac{W^0}{8}$ are the height and width. 
For the linguistic modality, we use the ``base-uncased'' version of BERT~\cite{devlin2018bert} as the text encoder to encode each word of the narrative caption to a real-valued vector, from which we extract features of noun phrases as $\bm{R}\in \mathbb{R}^{N\times C_{\text{r}}}$, where $N$ is the maximum number of noun phrases and $C_{\text{r}}$ is the channel number.

\subsection{Pixel-Phrase Matching Formulation}\label{sec:ppm}

For pixel-phrase matching, we directly use each noun phrase to group its corresponding pixels based on response values calculated between representations of all pixels and all noun phrases.
Concretely, we first project the visual features $\bm{F}$ and phrase features $\bm{R}$ to the same $C$-dimensional subspace by linear layers:
\begin{equation}
    \hat{\bm{F}}=\bm{F}\bm{W}_1, \hat{\bm{R}}=\bm{R}\bm{W}_2,
\end{equation}
where $\bm{W}_1\in \mathbb{R}^{C_{\text{v}}\times C}$ and $\bm{W}_2\in \mathbb{R}^{C_{\text{r}}\times C}$ are projection parameters, $\hat{\bm{F}}\in \mathbb{R}^{H\times W\times C}$ and $\hat{\bm{R}} \in \mathbb{R}^{N\times C}$ are projected features. 
Next, we first reshape $\hat{\bm{F}}$ to $\mathbb{R}^{HW\times C}$ and then conduct matrix multiplication between $\hat{\bm{F}}$ and $\hat{\bm{R}}$ to obtain response maps between all pixels and all noun phrases as follows:
\begin{equation}
    \bm{H}=\hat{\bm{R}}\hat{\bm{F}}^T,
\end{equation}
\begin{equation}
    \bm{M}=\sigma(\bm{H}),
\end{equation}
where $\hat{\bm{F}}^T\in \mathbb{R}^{C\times HW}$ is the transpose of $\hat{\bm{F}}$, $\bm{H} \in \mathbb{R}^{N\times HW}$ is the raw matching maps, and $\sigma$ denotes sigmoid function. 
After, we reshape $\bm{M}\in \mathbb{R}^{N\times HW}$ to $\mathbb{R}^{N\times H\times W}$ and $\bm{M}^n \in \mathbb{R}^{H\times W}$ is the response map of the $n$-th noun phrase, based on which we can generate the segmentation result.

It is straight-forward to train a pixel-phrase matching network: given ground-truth binary masks of all phrases $\bm{Y}\in \mathbb{R}^{N\times H\times W}$, we apply the binary cross-entropy (BCE) loss on the response maps $\bm{M}$. The operation can be written as:
\begin{equation}
    \mathcal{L}_{\text{bce}}(\bm{M}^{n,i},\bm{Y}^{n,i})=\bm{Y}^{n,i}\log(\bm{M}^{n,i})+(1-\bm{Y}^{n,i})\log(1-\bm{M}^{n,i}),
\end{equation}
\begin{equation}
    \mathcal{\overline{L}}_{\text{bce}}(\bm{M},\bm{Y})=-\frac{1}{NHW}\sum_{n=1}^{N}\sum_{i=1}^{HW}\mathcal{L}_{\text{bce}}(\bm{M}^{n,i},\bm{Y}^{n,i}).
    \label{eq:bce}
\end{equation}
It is worth noting that Eq.~\ref{eq:bce} can be rewritten as follows:
\begin{equation}
\begin{aligned}
    \mathcal{\overline{L}}_{\text{bce}}(\bm{M},\bm{Y})&=-\frac{1}{HW}\sum_{i=1}^{HW}\frac{1}{N}\sum_{n=1}^{N}\mathcal{L}_{\text{bce}}(\bm{M}^{n,i},\bm{Y}^{n,i}),\\
    &=-\frac{1}{HW}\sum_{i=1}^{HW}\mathcal{L}_{\text{multi-cls}}(\bm{M}^{:,i}, \bm{Y}^{:,i}),
\end{aligned}
\end{equation}
which shows that the pixel-phrase matching is equivalent to performing the multi-label classification process on each pixel if we consider $\bm{M}^{:,i}\in \mathbb{R}^{1\times N}$ as a probability distribution over all possible $N$ noun phrases (\textit{i.e.}, labels) for $i$-th pixel. In this view, it forces the model to produce high responses to positive phrases and otherwise to negative ones.

Since BCE loss treats each pixel separately, it can not handle the foreground-background sample imbalance problem. We apply Dice loss~\cite{milletari2016v} to alleviate this issue following previous works~\cite{zhang2021k,wang2020solov2,tian2020conditional}:
\begin{equation}
    \mathcal{L}_{\text{dice}}(\bm{M}^{n},\bm{Y}^{n})=1-\frac{2\sum_{i=1}^{HW}\bm{M}^{n,i}\bm{Y}^{n,i}}{\sum_{i=1}^{HW}\bm{M}^{n,i}+\sum_{i=1}^{HW}\bm{Y}^{n,i}},
\end{equation}
\begin{equation}
    \mathcal{\overline{L}}_{\text{dice}}(\bm{M},\bm{Y})=\frac{1}{N}\sum_{n=1}^N\mathcal{L}_{\text{dice}}(\bm{M}^{n},\bm{Y}^{n}).
\end{equation}

Overall, the final training loss function $\mathcal{L}_{\text{ppm}}$ of our PPMN can be formulated as:
\begin{equation}
    \mathcal{L}_{\text{ppm}}=\lambda_{\text{bce}}\mathcal{\overline{L}}_{\text{bce}}+\lambda_{\text{dice}}\mathcal{\overline{L}}_{\text{dice}},
\end{equation}
where $\lambda_{\text{bce}}$ and $\lambda_{\text{dice}}$ are hyperparameters used to balance these two losses. We empirically find $\lambda_{\text{bce}}=1$ and $\lambda_{\text{dice}}=1$ work best.

\subsection{Language-Compatible Pixel Aggregation}\label{sec:lcpa}

Without the explicit cross-modal interaction between visual and linguistic modalities, our model can only generate sub-optimal panoptic segmentation results using the limited category priors implied in noun phrases. 
To endow the phrase features with stronger discriminative ability, we propose a Language-Compatible Pixel Aggregation (LCPA) module to refine phrase features for multiple rounds.
This process is formulated as:
\begin{equation}
    \bm{R}^{l+1},\bm{H}^{l+1}=\text{LCPA}^l(\bm{F}, \bm{R}^{l}, \bm{H}^{l}), l=0,1,...,L-1,
\end{equation}
while $\bm{R}^{0}=\bm{R}$, $\bm{H}^{0}=\bm{H}$ and $L$ is the number of multiple rounds.

In the $l$-th round, we obtain indexes $\bm{H}_{\text{index}}^l\in \mathbb{R}^{N\times S\times 2}$ of the $S$ most compatible pixels for all phrases from $\bm{H}^l$:
\begin{equation}
    \bm{H}_{\text{index}}^l=\text{MaxPool}(\bm{H}^l, S),
\end{equation}
where $\text{MaxPool}(\cdot,S)$ is an adaptive max pooling layer that returns $S$ max pooling indexes instead of the values in our implementation. 
Afterwards, we use $\bm{H}_{\text{index}}^l$ to sample compatible pixel features $\bm{H}_{\text{f}}^l\in \mathbb{R}^{N\times S\times C}$ from $\bm{F}$.

\begin{figure*}[t]
   \centering
   \medskip
   \begin{subfigure}[t]{.33\linewidth}
  \centering
  \resizebox{0.95\linewidth}{!}{%
  \begin{tikzpicture}[/pgfplots/width=1.45\linewidth, /pgfplots/height=1.45\linewidth]
    \begin{axis}[% Axis labels
                 ymin=0,ymax=1,xmin=0,xmax=1,
    			 % Axis labels
        		 xlabel=IoU,
        		 ylabel=Recall@IoU,
         		 xlabel shift={-2pt},
        		 ylabel shift={-3pt},
         		 % General appearance
		         font=\small,
		         axis equal image=true,
		         enlargelimits=false,
		         clip=true,
		         % Grids
        	     grid style=solid, grid=both,
                 major grid style={white!85!black},
        		 minor grid style={white!95!black},
		 		 xtick={0,0.1,...,1.1},
                 xticklabels={0,.1,.2,.3,.4,.5,.6,.7,.8,.9,1},
        		 ytick={0,0.1,...,1.1},
                 yticklabels={0,.1,.2,.3,.4,.5,.6,.7,.8,.9,1},
         		 minor xtick={0,0.02,...,1},
		         minor ytick={0,0.02,...,1},
        		 % Legend
        		 legend style={at={(0.05,0.05)},
                 		       anchor=south west},
                 legend cell align={left}]
    \addplot+[red,solid,mark=none,ultra thick] table[x=IoU,y=ppm_overall]{figs/figure1_final.txt};
    \addlegendentry{PPMN}
    \addplot+[green,solid,mark=none,ultra thick] table[x=IoU,y=ppm_wococo_overall]{figs/figure1_final.txt};
    \addlegendentry{PPMN $\dag$}
    \addplot+[blue,solid,mark=none,ultra thick] table[x=IoU,y=PNG]{figs/figure1_final.txt};
    \addlegendentry{Gonz{\'a}lez~\textit{et al.}~\cite{gonzalez2021panoptic}}
    \end{axis}
\end{tikzpicture}}
  \subcaption{Overall performance}
  \label{fig:overall}
\end{subfigure}
\begin{subfigure}[t]{.33\linewidth}
  \centering
    \resizebox{0.95\linewidth}{!}{%
  \begin{tikzpicture}[/pgfplots/width=1.45\linewidth, /pgfplots/height=1.45\linewidth]
    \begin{axis}[% Axis labels
                 ymin=0,ymax=1,xmin=0,xmax=1,
    			 % Axis labels
        		 xlabel=IoU,
        		 ylabel=Recall@IoU,
         		 xlabel shift={-2pt},
        		 ylabel shift={-3pt},
         		 % General appearance
		         font=\small,
		         axis equal image=true,
		         enlargelimits=false,
		         clip=true,
		         % Grids
        	     grid style=solid, grid=both,
                 major grid style={white!85!black},
        		 minor grid style={white!95!black},
		 		 xtick={0,0.1,...,1.1},
                 xticklabels={0,.1,.2,.3,.4,.5,.6,.7,.8,.9,1},
        		 ytick={0,0.1,...,1.1},
                 yticklabels={0,.1,.2,.3,.4,.5,.6,.7,.8,.9,1},
         		 minor xtick={0,0.02,...,1},
		         minor ytick={0,0.02,...,1},
        		 % Legend
        		 legend style={at={(0.05,0.05)},
                 		       anchor=south west},
                 legend cell align={left}]
    \addplot+[red,dashed,mark=none,ultra thick] table[x=IoU,y=ppm_things]{figs/figure2_final.txt};
    \addlegendentry{PPMN (Things)}
    \addplot+[red,solid,mark=none,ultra thick] table[x=IoU,y=ppm_stuff]{figs/figure2_final.txt};
    \addlegendentry{PPMN (Stuff)}
    \addplot+[green,dashed,mark=none,ultra thick] table[x=IoU,y=ppm_wococo_things]{figs/figure2_final.txt};
    \addlegendentry{PPMN $\dag$ (Things)}
    \addplot+[green,solid,mark=none,ultra thick] table[x=IoU,y=ppm_wococo_stuff]{figs/figure2_final.txt};
    \addlegendentry{PPMN $\dag$ (Stuff)}
    \addplot+[olive,dashed,mark=none,ultra thick] table[x=IoU,y=MCN_things]{figs/figure2_final.txt};
    \addlegendentry{MCN~\cite{luo2020multi} (Things)}
    \addplot+[blue,dashed,mark=none,ultra thick] table[x=IoU,y=PNG_things]{figs/figure2_final.txt};
    \addlegendentry{Gonz{\'a}lez~\textit{et al.}~\cite{gonzalez2021panoptic} (Things)}
    \addplot+[blue,solid,mark=none,ultra thick] table[x=IoU,y=PNG_stuff]{figs/figure2_final.txt};
    \addlegendentry{Gonz{\'a}lez~\textit{et al.}~\cite{gonzalez2021panoptic} (Stuff)}
    \end{axis}
\end{tikzpicture}}
  \subcaption{Things and stuff categories}
  \label{fig:things_stuff}
\end{subfigure}
\begin{subfigure}[t]{.33\linewidth}
  \centering
      \resizebox{0.95\linewidth}{!}{%
  \begin{tikzpicture}[/pgfplots/width=1.45\linewidth, /pgfplots/height=1.45\linewidth]
    \begin{axis}[% Axis labels
                 ymin=0,ymax=1,xmin=0,xmax=1,
    			 % Axis labels
        		 xlabel=IoU,
        		 ylabel=Recall@IoU,
         		 xlabel shift={-2pt},
        		 ylabel shift={-3pt},
         		 % General appearance
		         font=\small,
		         axis equal image=true,
		         enlargelimits=false,
		         clip=true,
		         % Grids
        	     grid style=solid, grid=both,
                 major grid style={white!85!black},
        		 minor grid style={white!95!black},
		 		 xtick={0,0.1,...,1.1},
                 xticklabels={0,.1,.2,.3,.4,.5,.6,.7,.8,.9,1},
        		 ytick={0,0.1,...,1.1},
                 yticklabels={0,.1,.2,.3,.4,.5,.6,.7,.8,.9,1},
         		 minor xtick={0,0.02,...,1},
		         minor ytick={0,0.02,...,1},
        		 % Legend
        		 legend style={at={(0.05,0.05)},
                 		       anchor=south west},
                 legend cell align={left}]
    \addplot+[red,dashed,mark=none,ultra thick] table[x=IoU,y=ppm_singulars]{figs/figure3_final.txt};
    \addlegendentry{PPMN (Singulars)}
    \addplot+[red,solid,mark=none,ultra thick] table[x=IoU,y=ppm_plurals]{figs/figure3_final.txt};
    \addlegendentry{PPMN (Plurals)}
    \addplot+[green,dashed,mark=none,ultra thick] table[x=IoU,y=ppm_wococo_singulars]{figs/figure3_final.txt};
    \addlegendentry{PPMN $\dag$ (Singulars)}
    \addplot+[green,solid,mark=none,ultra thick] table[x=IoU,y=ppm_wococo_plurals]{figs/figure3_final.txt};
    \addlegendentry{PPMN $\dag$ (Plurals)}
    \addplot+[blue,dashed,mark=none,ultra thick] table[x=IoU,y=PNG_singulars]{figs/figure3_final.txt};
    \addlegendentry{Gonz{\'a}lez~\textit{et al.}~\cite{gonzalez2021panoptic} (Singulars)}
    \addplot+[blue,solid,mark=none,ultra thick] table[x=IoU,y=PNG_plurals]{figs/figure3_final.txt};
    \addlegendentry{Gonz{\'a}lez~\textit{et al.}~\cite{gonzalez2021panoptic} (Plurals)}
    \end{axis}
\end{tikzpicture}}
  \subcaption{Singulars and plurals}
  \label{fig:singulars_plurals}
\end{subfigure}
   \caption{\textbf{Average Recall Curve} for our PPMN method performance (a) compared to the state-of-the-art methods, and dissagregated into (b) things and stuff categories, and (c) singulars and plurals noun phrases.}
  \label{fig:average_recall_curve}
 \end{figure*}
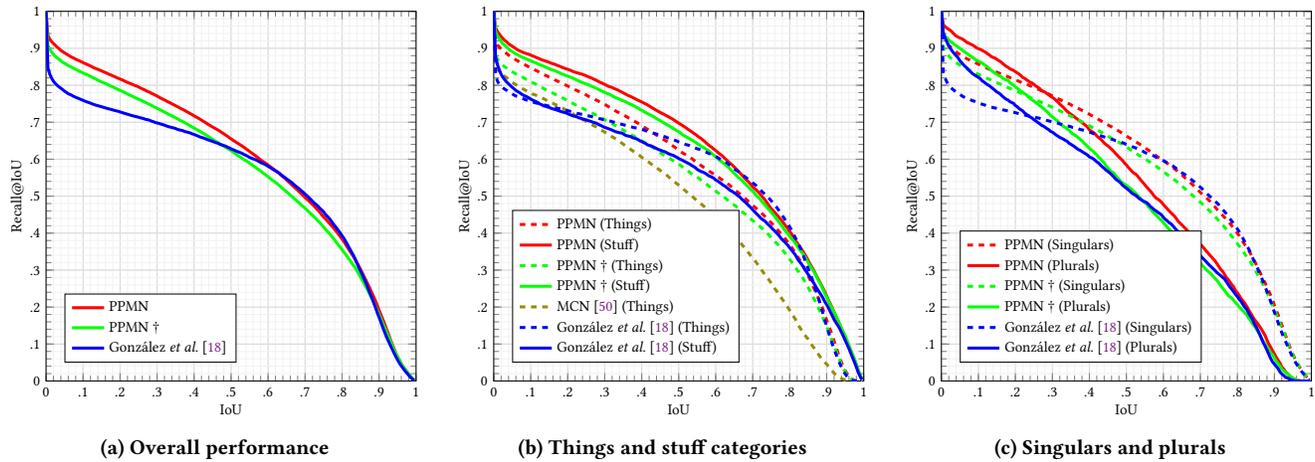
 
 \begin{table*}[t]
\caption{Comparison with state-of-the-art methods on the PNG benchmark, disaggregated into (a) things and stuff categories, and (b) singulars and plurals noun phrases. $\dag$ denotes training without COCO panoptic segmentation annotations.}
\label{tab:results}
\begin{subtable}{.5\linewidth}
  \centering
    \caption{Things and stuff categories.}
    \label{tab:things_stuff_results}
    {\tablestyle{5pt}{1}{
    \begin{tabular}{r||c|c|c}
    \hline\thickhline
    \rowcolor{mygray}
    & \multicolumn{3}{c}{Average Recall} \\
    \rowcolor{mygray}
    \multirow{-2}*{Method} & \multicolumn{1}{c}{overall} & \multicolumn{1}{c}{things} & \multicolumn{1}{c}{stuff} \\ \hline\hline
    Gonz{\'a}lez~\textit{et al.}~\cite{gonzalez2021panoptic} & 55.4 & 56.2 & 54.3 \\
    MCN~\cite{luo2020multi} & - & 48.2 & - \\ \hline\hline
    PPMN $\dag$ & 56.7 & 53.4 & 61.1 \\
    PPMN & \textbf{59.4 \color{sota_blue}{({+4.0})}} & \textbf{57.2 \color{sota_blue}{({+1.0})}} & \textbf{62.5 \color{sota_blue}{({+8.2})}} \\ \hline
    \end{tabular}
    }}
\end{subtable}%
\begin{subtable}{.5\linewidth}
  \centering
    \caption{Singulars and plurals noun phrases.}
    \label{tab:singulars_plurals_results}
  {\tablestyle{5pt}{1}{
    \begin{tabular}{r||c|c|c}
    \hline\thickhline
    \rowcolor{mygray}
    & \multicolumn{3}{c}{Average Recall} \\
    \rowcolor{mygray}
    \multirow{-2}*{Method} & \multicolumn{1}{c}{overall} & \multicolumn{1}{c}{singulars} & \multicolumn{1}{c}{plurals} \\ \hline\hline
    Gonz{\'a}lez~\textit{et al.}~\cite{gonzalez2021panoptic} & 55.4 & 56.2 & 48.8 \\ \hline\hline
    PPMN $\dag$ & 56.7 & 57.4 & 49.8 \\
    PPMN & \textbf{59.4 \color{sota_blue}{({+4.0})}} & \textbf{60.0 \color{sota_blue}{({+3.8})}} & \textbf{54.0 \color{sota_blue}{({+5.2})}} \\ \hline
    \end{tabular}
    }}
\end{subtable} 
\end{table*}

Next, we enhance the discriminative ability of each noun phrase by aggregating the visual context of its most compatible pixel features.
To this end, we follow the implementation practice of Transformer~\cite{vaswani2017attention} and revise it to a multi-head cross-modal attention mechanism $\text{MCA}(\cdot)$. Details of this process for the $n$-th noun phrase feature ${(\bm{R}^l)}^n\in \mathbb{R}^{1\times C}$ can be formulated as follows:
\begin{equation}
    \text{Attention}(\bm{Q},\bm{K},\bm{V})=\text{Softmax}(\frac{\bm{Q}\bm{W}_5{(\bm{K}\bm{W}_6)}^T}{\sqrt{C}})\bm{V}\bm{W}_7,
\end{equation}
\begin{equation}
    \text{MCA}=\mathbb{C}[{\{\text{Attention}((\bm{R}^l)^{n,d},(\bm{H}_{\text{f}}^l)^{n,d},(\bm{H}_{\text{f}}^l)^{n,d})\}}_{d=1}^D],
\end{equation}
where $\mathbb{C}(\cdot)$ denotes concatenation and $D$ is the number of heads.
$\bm{W}_5$, $\bm{W}_6$, and $\bm{W}_7$ are projection parameters.
$\bm{Q}$, $\bm{K}$, $\bm{V}$ are the query, key, and value respectively.
$(\bm{R}^l)^{n,d}\in \mathbb{R}^{1\times \frac{C}{d}}$ and $(\bm{H}^l_{\text{f}})^{n,d}\in \mathbb{R}^{1\times \frac{C}{d}}$ are input features of $d$-th head.
The output of $\text{MCA}(\cdot)$ is the refined $n$-th phrase feature, denoted as ${(\hat{\bm{R}}^l)}^n\in \mathbb{R}^{1\times C}$.
After applying $\text{MCA}(\cdot)$ on each phrase separately, we concatenate all refined phrase features and feed them to a standard Feed-Forward Network (FFN) to obtain the refined phrase features $\bm{R}^{l+1}\in \mathbb{R}^{N\times C}$:
\begin{equation}
    \hat{\bm{R}}^{l}=\mathbb{C}[{\{{(\hat{\bm{R}}^l)}^n\}}_{n=1}^N]+\bm{R}^l,
\end{equation}
\begin{equation}
    \bm{R}^{l+1}=\text{LN}(\text{FFN}(\text{LN}(\hat{\bm{R}}^l)))+\hat{\bm{R}}^l,
\end{equation}
where $\text{LN}$ is the LayerNorm~\cite{ba2016layer}.

At last, we project $\bm{F}$ and $\bm{R}^{l+1}$ to the same subspace.
Specifically, we project the visual features $\bm{F}$ with a $1\times 1$ convolution layer followed by GroupNorm~\cite{wu2018group} and ReLU activation~\cite{nair2010rectified}. 
As for phrase features, we apply a multi-layer perceptron (MLP) with 3 hidden layers to $\bm{R}^{l+1}$. 
Then we conduct matrix multiplication between them to acquire $\bm{H}^{l+1}\in \mathbb{R}^{N\times H\times W}$ and apply sigmoid function on $\bm{H}^{l+1}$ to obtain response maps $\bm{M}^{l+1}\in \mathbb{R}^{N\times H\times W}$. In the training phase, we apply $\mathcal{L}_{\text{ppm}}$ to all rounds of $\{\bm{M}^l\}_{l=0}^{L-1}$ for sufficient intermediate supervisions on the learning of our LCPA module:
\begin{equation}
    \mathcal{L}=\sum_{l=0}^{L-1}\mathcal{L}_{\text{ppm}}(\bm{M}^l,\bm{Y}).
\end{equation}
In the inference phase, we employ a threshold of 0.5 to obtain the grounding results from the last round response maps $\bm{M}^L$.

\section{Experiments}

\subsection{Dataset and Evaluation Criteria}
We evaluate our PPMN on the Panoptic Narrative Grounding benchmark~\cite{gonzalez2021panoptic}, which is extended from MS COCO~\cite{lin2014microsoft} dataset.
It includes 726,445 noun phrases from the whole Localized Narratives annotations~\cite{pont2020connecting} that are matched with 659,298 unique segments from MS COCO panoptic segmentation annotations. 
Each narrative caption contains an average of 11.3 noun phrases, of which 5.1 noun phrases are grounded.
Following Gonz{\'a}lez~\textit{et al.}~\cite{gonzalez2021panoptic}, we adopt the Average Recall to evaluate our PPMN.
Concretely, we first calculate the Intersection over Union (IoU) between segmentation predictions and ground-truth masks for all evaluated noun phrases.
Then, we compute recall at different IoU thresholds to obtain a curve where recall approaches one at very low IoU values and decreases at higher IoU values.
The Average Recall refers to the area under the curve described above. 
As for plural noun phrases, of which each phrase is annotated with multiple instances, we aggregate all corresponding ground truth masks into a single segmentation.

\subsection{Implementation Details}

Consistent with Gonz{\'a}lez~\textit{et al.}~\cite{gonzalez2021panoptic}, we use FPN~\cite{lin2017feature} with a ResNet-101~\cite{he2016deep} backbone pre-trained with Panoptic Feature Pyramid Network~\cite{kirillov2019panopticfpn} on MS COCO~\cite{lin2014microsoft} with 3x schedule using the official implementation~\cite{wu2019detectron2}. We fix parameters in FPN. The input image is resized with a shorter side to 800 and a longer side up to 1333, without changing the aspect ratio. For the linguistic input, we adopt the pretrained ``base-uncased'' BERT model~\cite{devlin2018bert} to convert each word in the narrative caption into a 768-dimensional vector. The maximum length of the input caption is 230, of which at most 30 different noun phrases need to be grounded. Adam~\cite{kingma2014adam} is utilized as the optimizer. We implement our proposed PPMN in PyTorch~\cite{paszke2019pytorch} and train it with batch size 12 for 14 epochs on 4 NVIDIA A100 GPUs. The initial learning rate is set to $1e^{-4}$, which is divided by 2 for every 2 epochs started from the 10-th epoch, while the learning rate for the text encoder is set to $1e^{-5}$ constantly. The default numbers of multiple rounds $L$ and language-compatible pixels $S$ are set to 3 and 200, respectively. In the inference phase, we average matching maps of all words included in each noun phrase following the setup of the two-stage baseline~\cite{gonzalez2021panoptic}.   

\subsection{Comparison with State-of-the-Art Methods}

We conduct experiments on the PNG benchmark to compare our PPMN with state-of-the-art methods. 
Since PNG is an emerging and challenging task, there are only two methods to compare at present.
Quantitative results are shown in Table~\ref{tab:results} and Figure~\ref{fig:average_recall_curve}.

Compared to the two-stage baseline~\cite{gonzalez2021panoptic}, our PPMN achieves significant performance boosts of 4.0/1.0/8.2/3.8/5.2 on the Average Recall evaluation metric for \textit{overall}/\textit{things}/\textit{stuff}/\textit{singulars}/\textit{plurals} splits (Table~\ref{tab:results}), indicating the superiority of our one-stage Pixel-Phrase Matching approach. 
From Figure~\ref{fig:average_recall_curve} we can observe that the area between the red (\textit{i.e.,} PPMN) and blue curves (\textit{i.e.,} Gonz{\'a}lez~\textit{et al.}~\cite{gonzalez2021panoptic}) is relative large when the IoU is less than 0.5, which means that our proposed PPMN can ground much more panoptic objects than the two-stage baseline.
Besides, our PPMN achieves comparable performances when the IoU is close to one, verifying that our PPMN can generate accurate and complete segmentation results without extra manually designed rules and post-processing.
As for the computational overhead, our pixel-phrase matching and LCPA module occupy only 1.14 GFLOPs and 2.03 GFLOPs with negligible parameters respectively, which are not computationally heavy.
To further demonstrate the effectiveness of our proposed PPMN, we directly use the ResNet-101~\cite{he2016deep} pretrained on ImageNet~\cite{krizhevsky2012imagenet} dataset as the image encoder, whose performances are shown in the row ``PPMN $\dag$'' of Table~\ref{tab:results} and the green curves in Figure~\ref{fig:average_recall_curve}. 
It can be seen that its Average Recall values exceed the two-stage baseline on all but one (\textit{i.e.,} \textit{things}) of splits, proving that our PPMN is strong enough to learn discriminative visual and linguistic features by end-to-end training, even without the panoptic priors obtained from pretraining on COCO panoptic annotations.
It is worth noting that the two-stage baseline is similar to a state-of-the-art phrase grounding method, \textit{i.e.,} GLIP~\cite{li2021grounded}, where they separately encode visual and phrase features and conduct cross-modal fusion before matching candidate regions with noun phrases.
Thus, the performance gap between our PPMN and the two-stage baseline indicates that directly adapting phrase grounding methods to PNG can not yield compelling results.
And we believe our PPMN can be a better baseline to promote research in PNG.

Multi-task Collaborative Network (MCN)~\cite{luo2020multi} is a state-of-the-art visual grounding method that achieves joint learning of REC and RES. 
Considering that REC and RES only include objects belonging to the \textit{things} category, we only evaluate its performance on the \textit{things} split of the PNG benchmark~\cite{gonzalez2021panoptic}.
As shown in Table~\ref{tab:things_stuff_results} and Figure~\ref{fig:things_stuff}, there is a big performance gap between MCN and methods designed for PNG.
We claim that this phenomenon is caused by the sparse and coarse annotations (\textit{i.e.,} box/segment-sentence pair in Figure~\ref{fig:teaser}) of previous grounding tasks.
In those settings, models are not forced to mine the fine-grained cross-modal semantic relationship between pixels and phrases, while our proposed method achieves this by the Pixel-Phrase Matching approach.

\subsection{Ablation Studies}

\begin{figure*}[t]
  \centering
  \includegraphics[width=0.95\linewidth]{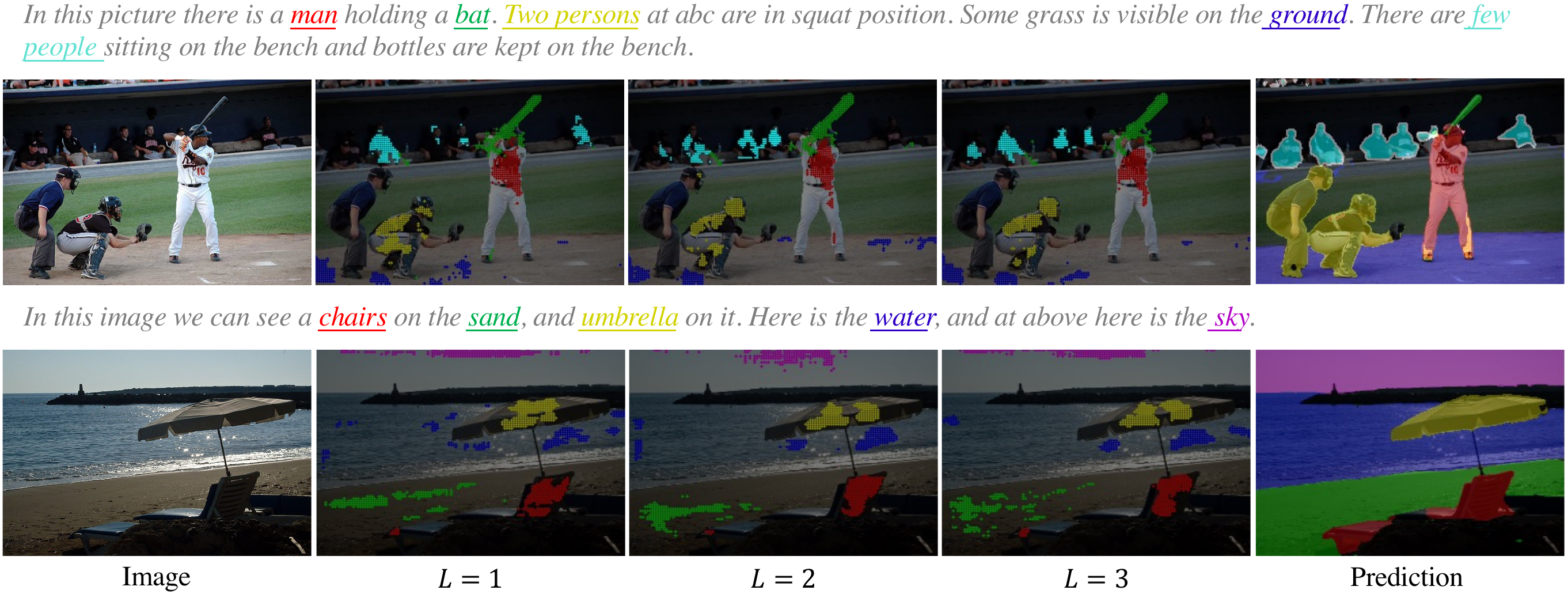}
  \caption{Visualization of the language-compatible pixels' locations in different rounds (2-nd to 4-th column) and segmentation results (5-th column).}
  \label{fig:topk}
\end{figure*}

To test if our LCPA module confers benefits, we conduct ablation studies on the PNG benchmark to evaluate its different designs.

\begin{table}[h]
    \caption{The number of language-compatible pixels $S$. We set the number of multiple rounds $L=1$ in this experiment.}
    \label{tab:ablation:language-compatible_pixels}
    \centering
    {
    \tablestyle{6pt}{1}\begin{tabular}{c||c|c|c|c|c}
    \hline\thickhline
    \rowcolor{mygray}
    & \multicolumn{5}{c}{Average Recall} \\
    \rowcolor{mygray}
    \multirow{-2}*{$S$} & \multicolumn{1}{c}{overall} & \multicolumn{1}{c}{singulars} & \multicolumn{1}{c}{plurals} & \multicolumn{1}{c}{things} & \multicolumn{1}{c}{stuff} \\ \hline\hline
    0 & 51.6 & 51.9 & 48.7 & 49.5 & 54.6 \\
    100 & 58.5 & 59.1 & 53.8 & 56.2 & 61.8 \\
    200 & \textbf{58.7} & \textbf{59.2} & \textbf{53.9} & \textbf{56.4} & \textbf{61.9} \\
    300 & 57.4 & 57.9 & 52.6 & 55.0 & 60.7 \\\hline
    \end{tabular}
    \vspace{-10pt}
    }

\end{table}

\paragraph{Number of Language-Compatible Pixels.} We evaluate different numbers of language-compatible pixels $S$ in the Table~\ref{tab:ablation:language-compatible_pixels}.
As shown in the 1-st and 2-nd rows, the performance boosts significantly when noun phrases are aware of their corresponding visual context, verifying that this cross-modal interaction can enhance the discriminative ability of noun phrase features. 
The best performance is achieved when $S=200$. Qualitative analysis of the language-compatible pixels in different rounds is shown in~\S\ref{sec:qualitative}.

\begin{table}[t]
    \caption{The number of multiple rounds $L$. We set the number of language-compatible pixels $S=200$ in this experiment.}
    \label{tab:ablation:iterative_rounds}
    \centering
    {
    \tablestyle{7pt}{1.14}\begin{tabular}{r||c|c|c|c|c}
    \hline\thickhline
    \rowcolor{mygray}
    & \multicolumn{5}{c}{Average Recall} \\
    \rowcolor{mygray}
    \multirow{-2}*{$L$} & \multicolumn{1}{c}{overall} & \multicolumn{1}{c}{singulars} & \multicolumn{1}{c}{plurals} & \multicolumn{1}{c}{things} & \multicolumn{1}{c}{stuff} \\ \hline\hline
    1 & 58.7 & 59.2 & 53.9 & 56.4 & 61.9 \\
    2 & 59.2 & 59.8 & 53.9 & 56.9 & 62.4 \\
    3 & 59.4 & 60.0 & \textbf{54.0} & \textbf{57.2} & 62.5 \\ 
    4 & \textbf{59.5} & \textbf{60.1} & \textbf{54.0} & 57.1 & \textbf{62.7} \\ \hline
    \end{tabular}
    }
    \vspace{-10pt}
\end{table}

\paragraph{Number of Multiple Rounds.} We demonstrate the influence of the number of multiple rounds $L$ in Table~\ref{tab:ablation:iterative_rounds}, 
where the Average Recall values on all evaluated splits grows gradually as $L$ increases. Considering the balance between computational overhead and performance, we choose $L=3$ in our final model.

\begin{figure*}[t]
  \centering
  \includegraphics[width=0.85\linewidth]{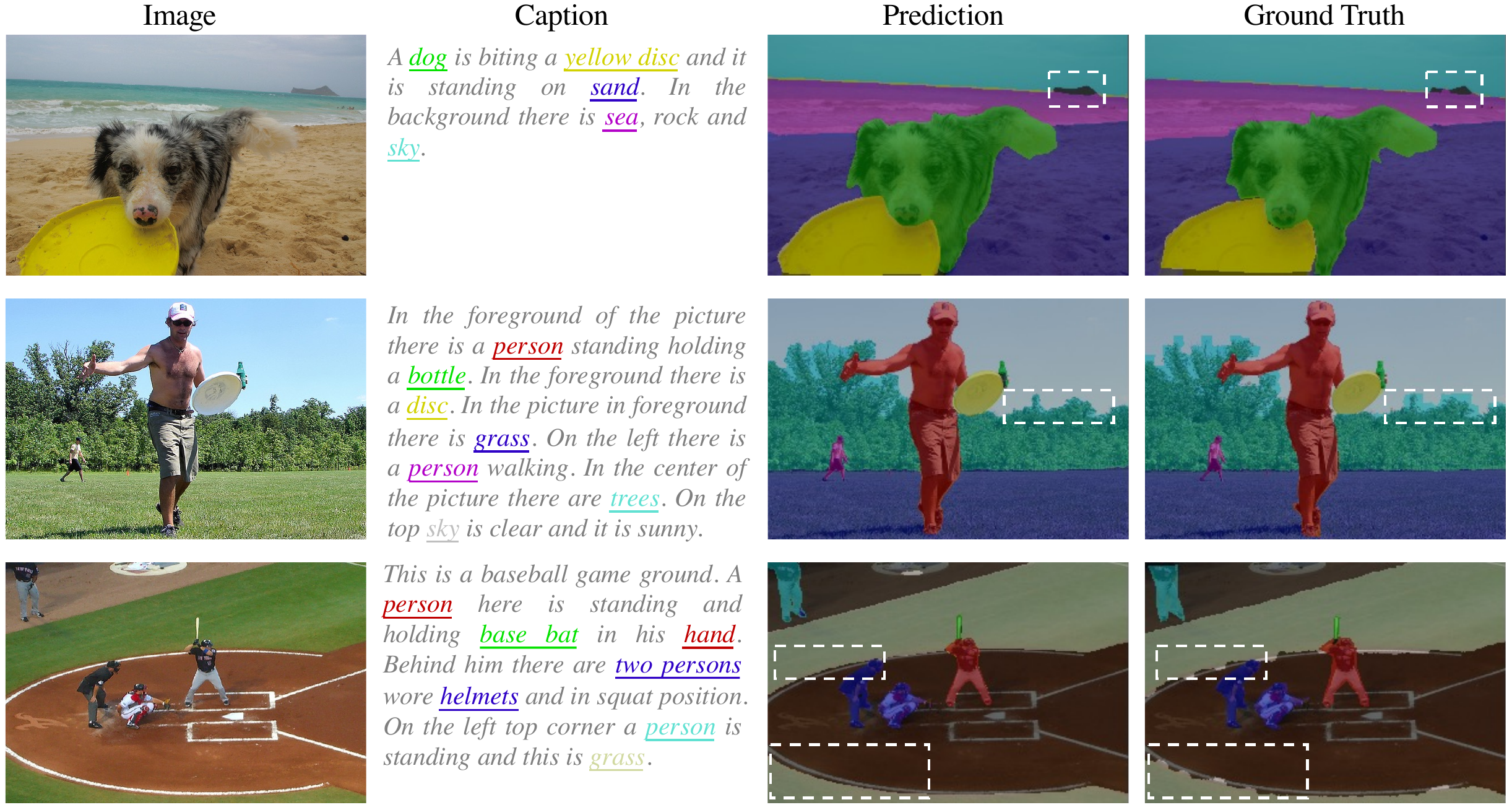}
  \caption{Qualitative analysis for our proposed PPMN. The regions where our PPMN produces better segmentation results than ground truth masks are highlighted within the white dashed boxes.}
  \label{fig:sota}
\end{figure*}

\paragraph{Visual Context Aggregation.} Here we demonstrate different ways of aggregating the visual context of the language-compatible pixels for each noun phrase (Table~\ref{tab:ablation:visual_context}).
MUTAN~\cite{ben2017mutan} is a conventional multi-modal fusion scheme originally designed for VQA~\cite{antol2015vqa}, which is able to model fine and rich cross-modal interactions between visual and language modalities based on a Tucker decomposition~\cite{tucker1966some}. 
Besides, we evaluate another implementation technique for the visual context aggregation by modifying the selective kernel convolution proposed in SKNet~\cite{li2019selective}.
Concretely, we first conduct average pooling on sampled compatible pixel features to obtain the integrated visual context for each noun phrase.
Then we fuse all noun phrase features and their corresponding visual context via element-wise multiplication and generate the channel-wise addition weights using a linear transformation.
After, we use these weights to adaptively fuse features of language-compatible pixels and noun phrases.
From Table~\ref{tab:ablation:visual_context} we can see that our LCPA is robust to the specific implementation techniques of the visual context aggregation part, and we use MCA (\S\ref{sec:lcpa}) in this paper for its best performances on all evaluated metrics.

\begin{table}[t]
    \caption{Different implementation techniques of the visual context aggregation part in LCPA.}
    \label{tab:ablation:visual_context}
    \centering
    {
    \tablestyle{7pt}{1.14}\begin{tabular}{r||c|c|c|c|c}
    \hline\thickhline
    \rowcolor{mygray}
    & \multicolumn{5}{c}{Average Recall} \\
    \rowcolor{mygray}
    \multirow{-2}*{Method} & \multicolumn{1}{c}{overall} & \multicolumn{1}{c}{singulars} & \multicolumn{1}{c}{plurals} & \multicolumn{1}{c}{things} & \multicolumn{1}{c}{stuff} \\ \hline\hline
    MUTAN~\cite{ben2017mutan} & 59.1 & 59.6 & 53.9 & 56.8 & 62.1 \\
    SKNet~\cite{li2019selective} & 59.2 & 59.8 & 53.8 & 56.8 & \textbf{62.5} \\
    MCA (\S\ref{sec:lcpa}) & \textbf{59.4} & \textbf{60.0} & \textbf{54.0} & \textbf{57.2} & \textbf{62.5} \\\hline
    \end{tabular}
    }
    \vspace{-10pt}
\end{table}

\subsection{Qualitative Analysis}\label{sec:qualitative}

In Figure~\ref{fig:topk}, we visualize locations of the language-compatible pixels in each round. 
Taking the 1-st row as an example, the language-compatible pixels for the plurals noun phrase ``few people'' (\textit{i.e.,} light blue points) can adaptively attend to regions of different people who are sitting on the chair in different rounds.
By this means, the noun phrase can aggregate sufficient visual context for all grounded people and segment them accurately and completely.

Figure~\ref{fig:sota} shows some qualitative results of our proposed PPMN compared to the ground truth masks. 
Surprisingly our proposed PPMN can produce better segmentation results than annotations.
In the 1-st row, our PPMN hardly generates false positive predictions on the rock when grounding the noun phrase ``sea''.
Furthermore, our PPMN segments finer and smoother boundaries between the trees and sky in the 2-nd row and between the grass and red soil field in the 3-rd row.

\section{Conclusion and Discussion}
In this paper, we propose a simple yet effective one-stage framework that can be end-to-end optimized for Panoptic Narrative Grounding (PNG), called Pixel-Phrase Matching Network (PPMN). 
The fine-grained Pixel-Phrase Matching strategy encourages our model to explore sufficient cross-modal semantic relevance from densely annotated pixel-phrase pairs of the PNG benchmark. 
Moreover, it can liberate our model from the performance bottleneck caused by the indirect two-stage baseline and cumbersome manually designed training/inference pipelines.
To further enhance the discriminative ability of noun phrase representations, we propose a Language-Compatible Pixel Aggregation (LCPA) module to adaptively aggregate representative visual context to each noun phrase from its corresponding language-compatible pixels for multiple rounds.
Experiments show that our PPMN outperforms previous methods by a large margin.
In the future, we plan to extend our PPMN to other visual grounding tasks, such as phrase grounding, to further verify its generalization ability.

\vspace{-5pt}

\section{Acknowledgments}

This work was supported in part by the National Natural Science Foundation of China under Grant 62122010 and Grant 61876177, in part by the Fundamental Research Funds for the Central Universities, and in part by the Key Research and Development Program of Zhejiang Province under Grant 2022C01082. 

\bibliographystyle{ACM-Reference-Format}
\balance
\bibliography{sample-base}

\end{document}